\documentclass[final,5p,times,twocolumn]{elsarticle}

\usepackage{amsmath,amssymb}
\usepackage{graphicx}
\usepackage{booktabs}
\usepackage{algorithm}
\usepackage{algpseudocode}
\usepackage{float}
\usepackage{hyperref}
\usepackage[mathscr]{eucal}
\usepackage{makecell} 
\usepackage{orcidlink}
\usepackage{comment}

\journal{Expert Systems with Applications}

\begin{document}

\begin{frontmatter}

\title{Policy-Based Reinforcement Learning with Action Masking for Dynamic Job Shop Scheduling under Uncertainty: Handling Random Arrivals and Machine Failures}

\author[inst1]{Sofiene Lassoued \corref{cor1}\orcidlink{0000-0001-7919-6939}}
\author[inst2]{Stefan Lier      \orcidlink{0000-0002-3314-7610}}
\author[inst1]{Andreas Schwung  \orcidlink{0000-0001-8405-0977}}

\cortext[cor1]{Corresponding author: Sofiene Lassoued, email: lassoued.sofiene@fh-swf.de}

\affiliation[inst1]{organization={South Westphalia University of Applied Sciences}, 
                    department={Automation Technology and Learning Systems}, 
                    addressline={Lübecker Ring 2}, 
                    city={Soest}, 
                    postcode={59494}, 
                    state={North Rhine-Westphalia}, 
                    country={Germany}}

\affiliation[inst2]{organization={South Westphalia University of Applied Sciences}, 
                    department={Logistik und Supply Chain Management}, 
                    addressline={Lindenstr.53}, 
                    city={Meschede}, 
                    postcode={59872}, 
                    state={North Rhine-Westphalia}, 
                    country={Germany}}

\begin{abstract}
We present a novel framework for solving Dynamic Job Shop Scheduling Problems under uncertainty, addressing the challenges introduced by stochastic job arrivals and unexpected machine breakdowns. Our approach follows a model-based paradigm, using Coloured Timed Petri Nets to represent the scheduling environment, and Maskable Proximal Policy Optimization to enable dynamic decision-making while restricting the agent to feasible actions at each decision point. To simulate realistic industrial conditions, dynamic job arrivals are modeled using a Gamma distribution, which captures complex temporal patterns such as bursts, clustering, and fluctuating workloads. Machine failures are modeled using a Weibull distribution to represent age-dependent degradation and wear-out dynamics. These stochastic models enable the framework to reflect real-world manufacturing scenarios better. In addition, we study two action-masking strategies: a non-gradient approach that overrides the probabilities of invalid actions, and a gradient-based approach that assigns negative gradients to invalid actions within the policy network. We conduct extensive experiments on dynamic JSSP benchmarks, demonstrating that our method consistently outperforms traditional heuristic and rule-based approaches in terms of makespan minimization. The results highlight the strength of combining interpretable Petri-net-based models with adaptive reinforcement learning policies, yielding a resilient, scalable, and explainable framework for real-time scheduling in dynamic and uncertain manufacturing environments.

\end{abstract}


\end{frontmatter}

\section{Introduction}
\label{sec: Introduction}

In today's competitive market, adapting to changes and overcoming disruption is essential. At the core of industrial operations lies the scheduling of manufacturing lines, a critical component that directly affects productivity and responsiveness. Scheduling problems are traditionally modeled as Job Shop Scheduling Problems (JSSP), which focus on optimizing the allocation of jobs to machines under fixed conditions. However, these classical models fail to account for real-world disruptions such as material shortages, machine breakdowns, or fluctuating resource availability. This limitation highlights the need for more resilient approaches, such as Dynamic Job Shop Scheduling (DJSS), which provides flexibility to handle unforeseen events through strategies including reactive, predictive-reactive, and robust proactive scheduling \cite{Tarek.2014}.

Solving scheduling problems has traditionally relied on three families of optimization methods: exact solutions, heuristics/metaheuristics, and, more recently, iterative AI-based approaches. Exact methods, such as branch and bound \cite{Brucker.1994} or linear Programming algorithms \cite{Floudas.2005}, guarantee optimal solutions when they exist. However, since these methods must explore the entire solution space, their computational cost becomes prohibitive for larger problems, where the number of possible states grows exponentially, as is typical for NP-hard problems like the JSSP \cite{Garey.1976}. As a result, exact methods can become infeasible for complex or large-scale scheduling tasks. Researchers have sought a compromise by accepting suboptimal solutions in exchange for faster response times, using heuristics and metaheuristics to overcome these limitations.

Heuristics solve scheduling problems by applying a set of predefined rules or algorithms tailored to specific cases \cite{Nohair.2022}. However, heuristics are often problem-specific and require deep domain expertise to develop effective rules. This is where metaheuristics come into play, strategies inspired by natural processes such as ant colony optimization  \cite{Engin.2018}, genetic algorithms  \cite{Yu.2018}, and swarm intelligence and Evolutionary Algorithms \cite{Gao.2019} offer a promising alternative. Metaheuristics excel at efficiently exploring large search spaces and finding good solutions, but they face two significant challenges: they require extensive parameter tuning, which limits their generalization, and they usually lack explicit knowledge retention. This means that if the problem parameters change, the algorithm must often restart the search process from scratch, reducing its effectiveness in dynamic environments.

In recent years, AI-based solutions, particularly those leveraging reinforcement learning (RL), have demonstrated significant potential in tackling the challenges of dynamic and uncertain scheduling environments, where adaptability is essential \cite{Panzer.2022}. RL methods generally fall into three categories: value-based, policy-based, and actor-critic algorithms. Pure value-based approaches, such as Deep Q-Networks (DQN) \cite{Zhou.2021}, estimate action-value functions to guide decision-making. Pure policy-based methods, like REINFORCE \cite{Wu.2020}, optimize policies directly using Monte Carlo estimates but often suffer from high variance in their updates. Hybrid actor-critic algorithms, including Proximal Policy Optimization (PPO) \cite{Park.2021} and Advantage Actor-Critic (A2C) \cite{Hubbs.2020}, combine value estimation and policy optimization, providing a balance that has proven effective across a wide range of scheduling problems.

While RL algorithms have advanced significantly, the environment, a critical component of the RL framework, has received comparatively less attention. Traditional scheduling models are not inherently designed for RL integration, often requiring complex reformulations into structures like disjunctive graphs. These are then encoded using methods such as Graph Neural Networks \cite{Hameed.2023} or Transformers \cite{Li.2024} to create structured observations suitable for learning. In addition, such environments overlook the highly constrained nature of JSSPs, where only a small subset of actions is valid at any time. Consequently, RL agents waste valuable training samples on invalid actions, reducing learning efficiency and slowing convergence. In addition, in the limited cases where the action masking is applied directly on the policy network output layer logits\cite{Huang.2022}, while effective, the agent will not learn the masking behaviour and will still depend on an external mask  to be provided even after training.

Effectively solving complex scheduling problems requires not only a capable RL agent but also an environment that reflects real-world dynamics and enables seamless interaction. We extend our previous PetriRL framework for JSSPs \cite{Lassoued.2024} to accommodate dynamic events such as random job arrivals and machine breakdowns. This approach combines the formal structure of Petri nets with the adaptability of RL. The Petri net is the backbone of the simulator: transitions represent the RL agent's action space, markings serve as observations, and guard functions provide dynamic action masking to enforce feasibility. This integration enhances sample efficiency through masking, improves interpretability via token flow, and supports a modular, scalable design by leveraging the inherent composability of Petri nets.

Ultimately, this work addresses a key challenge in dynamic scheduling: the lack of a unified framework for modeling and benchmarking. To support reproducible research, we provide an open-source Python package, \textbf{PetriRL} (Gym compatible and available on PyPI), which enables consistent simulation, fair comparisons, and a streamlined focus on optimization algorithm development.

\bigskip
The contributions of this study can be summarized as follows.
\begin{enumerate}

    \item We propose a dynamic job shop scheduling framework that integrates CPTN with MPPO RL, enabling adaptive scheduling under uncertainties such as random job arrivals and machine breakdowns.

    \item We put to the test two invalid action-making concepts, a gradient-free action masking to apply external constraints like machine breakdown, and a gradient-based approach, to approximate the guard function of the petrinet and enforcing the DJJP constains.

    \item Our framework models random job arrivals and machine breakdowns explicitly within the CTPN environment, enabling realistic simulation of dynamic manufacturing scenarios and real-time adaptive decision-making by the RL agent.

    \item We validate the proposed approach on benchmark dynamic JSSP problems, demonstrating improvements in makespan minimization and fault tolerance compared to various dispatching rules heuristics.

    \item We conduct ablation studies to quantify the impact of the different elements of our approach in maintaining robustness and efficiency under stochastic disruptions.
\end{enumerate}
   
\bigskip
This paper is structured as follows: Section \ref{sec: related_work} reviews related literature and highlights its limitations to identify the research gap. Section \ref{sec: background and preliminaries} provides background on the elements used in our approach, namely the RL framework and timed colored Petri nets. 

In Section \ref{sec: problem_formulation}, we provide definitions and details regarding the JSSP and DJSSP. Section \ref{sec: dynamic_fault_tolerant} outlines our methodology, beginning with presenting the JSSP RL's environment model using timed colored petri nets. This section also includes modeling random operation time arrivals and machine downtimes, and masking invalid actions. 

In Section \ref{sec: Results and Discussion}, we present our findings, starting with selecting benchmarks, followed by the experimental setup and training performance. The results are presented incrementally, first focusing on performance with only machine breakdown scenarios, then a combination of machine breakdown scenarios and random job arrival. This section concludes with an ablation study. 

Finally, Section \ref{sec: Conclusion} summarizes the key insights gained from this study and suggests promising avenues for future research.

\section{Related Work}
\label{sec: related_work}

Dynamic scheduling under uncertainty has been extensively studied, with various approaches developed to handle challenges such as random job arrivals and machine breakdowns. This review first presents established scheduling approaches and strategies, then narrows its focus to RL-based methods to address dynamic scheduling challenges under uncertainty.

\begin{figure*}[ht]
\centering
\includegraphics[width=\linewidth]{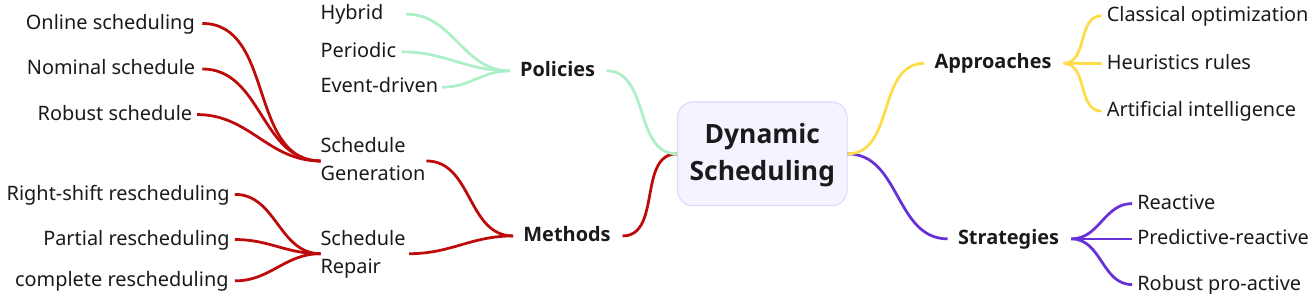}
\caption{A taxonomy map of dynamic scheduling, highlighting the main branches, approaches, strategies, policies, and methods~\cite{Wang.2020}.}
\label{fig:taxonomie}
\end{figure*}

The Dynamic Job Shop Scheduling Problem (Dynamic JSSP) extends the classical JSSP by incorporating real-time events that capture uncertainties \cite{Wang.2020}. These events can be broadly classified into two categories: resource-related and job-related \cite{Stoop.1996} \cite{Vieira.2003}. Resource-related disruptions include machine breakdowns, tool failures, and operator absences. Job-related events encompass job cancellations, due date changes, and priority updates. Addressing Dynamic JSP requires adaptive rescheduling techniques that can respond promptly to disruptions, maintaining efficient and reliable production. In Figure \ref{fig:taxonomie}, we present a taxonomy to help navigate the various subjects and subtopics within dynamic scheduling. The framework primarily focuses on four key subjects: Policies, Approaches, Strategies, and Methods.

At the policy level, three main policies are commonly adopted in dynamic scheduling: periodic rescheduling, event-based, and hybrid. The authors of \cite{Jin.2017} addressed integrated process planning and scheduling (IPPS) with new job arrivals using MILP (Mixed Integer Linear Programming). They found that event-driven scheduling greatly reduces computational effort without compromising performance, especially in systems with rare but significant events. However, purely event-driven methods may overlook gradual changes. The hybrid approach, which combines periodic checks with event triggers, offers the optimal balance between efficiency, safety, and responsiveness.

At the strategic level, scheduling approaches are typically classified as reactive, predictive-reactive, and robust-proactive. Robust scheduling anticipates manufacturing disturbances; for example, \cite{Grumbach.2024} applied DRL to generate proactive baseline schedules that absorb uncertainties by adjusting operation slack times through stretching or compressing plan durations. Predictive-reactive scheduling creates an initial baseline schedule and updates it in response to real-time events; \cite{Pleier.2024} employed a two-stage heuristic combined with simulation to enhance robustness by integrating pre-planning with real-time adjustments. Finally, completely reactive scheduling (also known as online scheduling) generates no pre-schedule and makes decisions on the fly. \cite{Ren.2024} Uses a Dueling Double DQN (D3QN) agent to choose dispatch rules dynamically upon event triggers.

Methods for generating schedules in DJSSPs fall into two main categories: creating complete schedules upfront or repairing existing ones, often aligned with the chosen strategy. Predictive-reactive strategies rely on schedule repair, while robust proactive approaches generate fully robust schedules from the start \cite{Ouelhadj.2009}. For example, \cite{He.2013} applies right-shift rescheduling and route changes in a predictive-reactive framework to handle machine breakdowns in flexible job shops.

At the approach level, DJSSPs have been tackled using exact methods, heuristics/metaheuristics, and recently AI-based techniques like DRL. For example, \cite{Lunardi.2020} addressed an online printing operation using two exact methods, namely MILP and Constraint Programming (CP). MILP was only able to solve small instances due to its high computational cost, whereas CP scaled better, benefiting from commercial solvers that combine exact and heuristic methods. applied an evolutionary genetic algorithm to minimize makespan, showing promising results in dynamic settings. Building on the success of RL combined with deep neural networks in AlphaGo \cite{Silver.2017}, Deep Reinforcement Learning (DRL) has increasingly been applied to DJSSPs for its adaptability to stochastic and evolving environments, as reviewed by \cite{Ngwu.2025}. The following paragraphs will explore the use of RL in solving DJSSP in more detail.

%%---------------------------- Focus paper DRL  --------------------------

The authors \cite{Hu.2020} modeled flexible manufacturing systems using timed S3PR  to capture operation durations, resource constraints, and deadlock behavior. They propose a specialized graph convolutional network layer called the Petri Net Convolution (PNC) layer, to encode the Petri net state, which is then processed by a Deep Q-Network (DQN) to learn a scheduling policy. While effective, the method relies on fixed system structures, limiting its generalizability to more complex or modular environments. Instead of a value-based approach like DQN, the authors \cite{Zhang.2022} used a policy-based method, namely Proximal Policy Optimization (PPO), as a dynamic scheduling method. Although the approach differentiates between valid and invalid actions, masking is not applied directly within the policy network. Instead, invalid actions are filtered out after the policy selects an action, which can lead to wasted computation and potentially suboptimal decisions. \cite{Liu.2023} presents a size-agnostic DRL based on graph neural networks and PPO to address the DJSSP with stochastic job arrivals and machine breakdowns. By modeling jobs and machines as a disjunctive graph, the framework enables real-time scheduling decisions and demonstrates generalization across different problem sizes without requiring retraining.

%%---------------------------- Actions masking paper DRL  -

In large, highly constrained action spaces, invalid-action masking becomes essential \cite{Ye.2020}. The authors in \cite{Huang.2022} compared four invalid-action elimination strategies: no masking as a control baseline, penalizing invalid action with penalties in the reward function, naïve action removal, and proper logit-level masking. Penalty methods \cite{Dietterich.2000} introduce negative rewards but scale poorly. Naïve masking eliminates invalid actions during sampling but leaves the policy unchanged, causing their probabilities to decay only through vanishing gradients and increasing the KL divergence between successive policies. Proper masking instead modifies logits directly by assigning large negative values to invalid actions before the softmax, reliably forcing their probabilities to zero. The results show that action removal only works in small action spaces, while penalty-based schemes degrade severely as dimensionality grows and require delicate reward tuning. Based on the work of \cite{Huang.2022}, a core limitation identified by the authors of  \cite{} is that none of these approaches learn the masking itself: the agent depends entirely on an externally provided mask. To address this, the authors introduce three methods: Off-PIAM, On-PIAM, and CO-IAM, each using two policy networks: a masked policy for safe trajectory generation and a raw policy that still assigns probability to invalid actions. Off-PIAM updates the raw policy via importance sampling IS; On-PIAM trains only the masked policy; and CO-IAM couples both via a composite loss.

%%-------------------------------------- Current  Limitations -----------------

\bigskip

Research on DRL for the Dynamic JSSP still faces several gaps. There is no standardized benchmark or evaluation protocol, making comparisons across studies inconsistent. Disruptions are often modeled with simplistic probability distributions, and few works address multiple uncertainties—such as machine failures and job arrivals—at the same time. The action space also remains problematic: many approaches filter invalid actions only after the policy outputs them, wasting computation and destabilizing learning. Existing masking techniques generally rely on externally provided masks, meaning the agent does not learn the underlying feasibility structure. The CO-IAM came the closesd to adress the masking problem, but requires two separate policies and uses advantage to

\section{Background and Preliminaries}
\label{sec: background and preliminaries}

In this chapter, we introduce the key components of our approach, namely the Timed Colored Petri Net and the Reinforcement Learning framework.

\subsection{Reinforcement Learning}
\label{subsec: RL Preliminaries}
Reinforcement Learning (RL) is a machine learning paradigm in which agents learn to make sequential decisions by interacting with an environment to maximize cumulative rewards. RL is formalized using a Markov Decision Process (MDP) defined by the tuple $(\mathcal{S}, \mathcal{A}, \mathcal{P}, \mathcal{R}, \gamma)$, where $\mathcal{S}$ is the state space, $\mathcal{A}$ the action space, $\mathcal{P}(s' |s, a)$ the state transition probability, $\mathcal{R}(s, a,s')$ the reward function, and $\gamma \in [0,1)$ the discount factor. The objective is to learn a policy $\pi^*$ that maximizes the expected discounted return:

\begin{equation}
\label{eq:return}
G_t = \sum_{k=0}^{\infty} \gamma^{k} R_{t+k+1}.
\end{equation}
\vspace{1em}

RL algorithms fall into three main categories: value-based, policy-based, and hybrid methods \cite{Richard.1998}. Value-based methods estimate the expected return using a state-value function $V(s)$ or a state-action function $Q(s, a)$, typically combined with an exploration policy such as $\epsilon$-greedy. Policy-based methods directly optimize the policy without relying on value estimates. Hybrid methods, such as actor-critic, combine both approaches by using value estimates to guide policy updates.

In this work, we adopt Maskable Proximal Policy Optimization (MPPO) \cite{RomanBartak.2022}, a variant of PPO \cite{Schulman.2017} that integrates action masking. PPO introduces a clipping mechanism to constrain policy updates, enhancing training stability. The maskable extension enables dynamic pruning of invalid actions via Boolean masks, derived in our case from Petri net guard functions. This enforces domain constraints and improves learning efficiency, especially beneficial in NP-hard problems like the JSSP.

\subsection{Colored-Timed Petrinets}
\label{subsec: colored-Timed Petrinets}

As introduced in Section~\ref{subsec: RL Preliminaries}, RL involves an agent interacting with an environment, either real or simulated, to learn optimal policies. In this study, we adopt a model-based RL (MBRL) approach by modeling the environment using a Petri Net. This modeling choice offers several MBRL advantages: (i) improved sample efficiency, since learning can occur with fewer environment interactions; (ii) enhanced interpretability through explicit representations of concurrency, resource constraints, and state transitions; and (iii) the ability to conduct controlled simulations without relying on costly real-world experiments.

Petri Nets are formal graphical and mathematical tools used to model discrete-event systems characterized by concurrency, synchronization, and resource sharing. A Petri Net is defined as a pair $(\mathcal{G}, \mu_0)$, where $\mathcal{G}$ is a bipartite graph consisting of places $\mathcal{P}$ and transitions $\mathcal{T}$, and $\mu_0$ is the initial marking, denoting the token distribution over places. Tokens abstract resources or job states and flow through the net by firing transitions, thereby reflecting the system's dynamics.

Each node $n \in \mathcal{P} \cup \mathcal{T}$ has a set of input nodes $\pi(n)$ and output nodes $\sigma(n)$. A transition $t \in \mathcal{T}$ is enabled when each input place $p \in \pi(t)$ contains at least one token. Upon firing, the new marking $\Tilde{\mu}$ is defined as:

\begin{equation}
\Tilde \mu(p)=
    \begin{cases}
     \mu(p)-1 & \forall\ p \in \pi(t) \\
     \mu(p)+1 & \forall\ p \in \sigma(t) \\
     \mu(p) & \text{otherwise}
    \end{cases}  
    \label{eq: arc expression }
\end{equation}

\vspace{1em}

\begin{figure*}[ht]
\centering
    \includegraphics[width=1\linewidth]{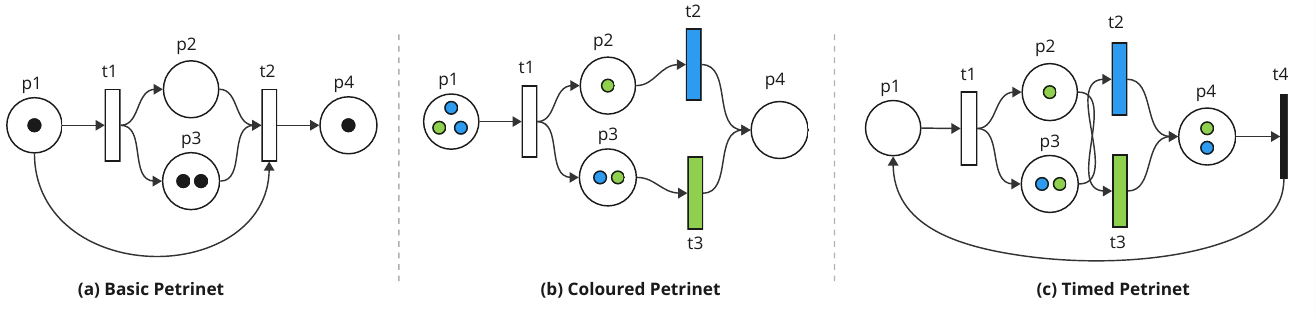}
\caption{Examples of Petri Net Variants: (a) Basic Petri Net, (b) Colored Petri Net, and (c) Colored-Timed Petri Net.}
\label{fig: Petri Nets example}
\end{figure*}

In addition to basic transitions, timed transitions can model process delays by enforcing a minimum sojourn time for tokens in input places. This is essential for representing machine processing times or transportation delays.

Colored Petri Nets (CPNs) extend this formalism by associating additional information with tokens called colors, allowing compact modeling of systems with repeated structures but varying attributes. This makes CPNs particularly effective for modeling job-shop scheduling problems with heterogeneous jobs~\cite{Jensen.1991}. A CPN is formally defined as:

\begin{equation}
\label{eq:cpn_definition}
\text{CPN} = (\mathcal{P}, \mathcal{T}, \mathcal{A}, \Sigma, \mathit{C}, \mathit{N}, \mathit{E}, \mathit{G}, \mathit{I})
\end{equation}
Where:
\begin{align*}
\mathcal{P} \quad &\text{: set of places} \\
\mathcal{T} \quad &\text{: set of transitions} \\
\mathcal{A} \quad &\text{: set of arcs} \\
\Sigma \quad &\text{: set of token colors} \\
\mathit{C} \quad &\text{: color set , } \mathit{C}: \mathcal{P} \cup \mathcal{T} \rightarrow \Phi(\Sigma) \\
\mathit{N} \quad &\text{: arc direction, } \mathit{N}: \mathcal{A} \rightarrow (\mathcal{P} \times \mathcal{T}) \cup (\mathcal{T} \times \mathcal{P}) \\
\mathit{E} \quad &\text{: arc expression, } \mathit{E}: \mathcal{A} \rightarrow e \\
\mathit{G} \quad &\text{: guard function, } \mathit{G}: \mathcal{T} \rightarrow \{0, 1\} \\
\mathit{I} \quad &\text{: initialization , } \mathit{I}: \mathcal{P} \rightarrow \text{init sequence}
\end{align*}

Figure~\ref{fig: Petri Nets example} illustrates three Petri Net variants: (a) a basic Petri Net with mono-color tokens logic; (b) a Colored Petri Net where tokens carry attributes; and (c) a Colored-Timed Petri Net where transitions can include timing constraints.

In (a), transition $t_1$ is enabled because its input place holds a token, while $t_2$ is not since $p_2$ is empty. In (b), although $p_2$ contains a token, $t_2$ is not enabled because the token's color does not satisfy the transition's guard. In (c), a timed transition fires only after the token's sojourn time in $p_4$ has elapsed.

\section{Problem Formulation and Modelling}
\label{sec: problem_formulation}

\subsection{Job Shop Scheduling Problem}

The Job Shop Scheduling Problem consists of a set of \(n\) jobs \(\mathcal{J} = \{1, 2, \dots, n\}\) and \(m\) machines \(\mathcal{M} = \{1, 2, \dots, m\}\). Each job \(j \in \mathcal{J}\) is defined as an ordered sequence of operations \(O_{j1}, O_{j2}, \dots, O_{j,\ell_j}\). Each operation \(O_{jk}\) must be processed on a specific machine \(M_{jk} \in \mathcal{M}\) for a fixed processing time \(p_{jk} > 0\).

The goal is to minimize the makespan \(C_{\max}\), the maximum completion time across all jobs, by scheduling operation start times \(S_{jk}\), where \(j \in \mathcal{J}\) indexes jobs and \(k\) indexes operations within job \(j\).

The problem is subject to the following constraints: (1) job precedence, ensuring operations follow the job order; (2) the makespan definition, linking \(C_{\max}\) to operation completion times; and (3) machine conflict, preventing overlapping operations on the same machine;

\begin{equation*}
\begin{aligned}
    &\text{(1)}\quad S_{j,k+1} \geq S_{jk} + p_{jk}, \quad \forall j \in \mathcal{J},\; k < \ell_j, \\
    &\text{(2)}\quad C_{\max} \geq S_{jk} + p_{jk}, \quad\forall j \in \mathcal{J},\; k \leq \ell_j.\\
    &\text{(3)}\quad S_{jk} \geq S_{j'k'} + p_{j'k'} \;\text{or}\; S_{j'k'} \geq S_{jk} + p_{jk}, \\
    & \quad\quad \forall (j,k) \neq (j',k'),\; M_{jk} = M_{j'k'}, \\
\end{aligned}
\end{equation*}

The Formulation provided above models a traditional JSSP Problem. To incorporate machines breakdowns, we extend the model by introducing a set of breakdown intervals \(\mathcal{B}_m = \{[b_{m1}^{\text{start}}, b_{m1}^{\text{end}}], \dots\}\) for each machine \(m \in \mathcal{M}\). During these intervals, machine \(m\) is unavailable for job processing. Two key assumptions are made:  
(4) No new operations can start on a machine during a breakdown;  
(5) Ongoing operations are paused during a breakdown and resume once the machine becomes available, thus extending their effective processing time.

Let \(C_{jk}\) denote the actual completion time of operation \(O_{jk}\), and let \(P_{eff}(jk)\) denote the effective processing time, which includes any delay due to breakdowns overlapping with the operation window. The modified constraints become:

\begin{equation*}
\begin{aligned}
    &\text{(4)}\quad S_{jk} \notin \bigcup \mathcal{B}_{M_{jk}}, \quad \forall j,k, \\
    &\text{(5)}\quad C_{jk} = S_{jk} + P_{eff}(jk), \quad \forall j,k.
\end{aligned}
\end{equation*}

The effective processing time \(P_{eff}(jk)\) is computed as:

\begin{equation*}
P_{eff}(jk) = p_{jk} + \sum_{\substack{[b^{\text{start}}, b^{\text{end}}] \in \mathcal{B}_{M_{jk}} \\ 
[b^{\text{start}}, b^{\text{end}}] \cap [S_{jk}, S_{jk}+p_{jk}] \neq \emptyset}} 
(b^{\text{end}} - b^{\text{start}})
\tag{6}
\end{equation*}

\subsection{Gradient-free action masking}

A common method for handling invalid actions in reinforcement learning is logit-level \emph{masking}, which assigns large negative values to the logits of invalid actions, effectively reducing their softmax probabilities to zero. Because this technique does not directly influence the policy through gradients, it is referred to in this manuscript as "gradient-free."

For an action $a_i$ with logit $z_i$, the softmax probability is given by
\begin{equation}
\pi_i = \frac{e^{z_i}}{\sum_{j=1}^{n} e^{z_j}},
\end{equation}

Suppose an action $a_k$ is invalid then :

\begin{equation}
\pi_k = \frac{e^{-M}}{\sum_{j=1}^{n} e^{z_j}} \approx 0. \quad M \gg 0
\end{equation}

For valid actions $a_i$ ($i \neq k$), the softmax effectively normalizes only over valid logits:

\begin{equation}
\pi_i \approx \frac{e^{z_i}}{\sum_{j \neq k} e^{z_j}}.
\end{equation}
  
After selecting an action using a sampling or greedy strategy and computing the corresponding advantage $\hat{A}_t$, the policy network calculates the gradient to update its weights. Initially, this gradient is computed at the softmax output layer and then backpropagated through the rest of the network. The gradient of the log-softmax with respect to logits $z_j$ is
\begin{equation}
\frac{\partial \log \pi_i}{\partial z_j} =
\begin{cases}
1 - \pi_i, & i = j, \\
- \pi_j, & i \neq j,
\end{cases}
\end{equation}
so that for the sampled action $a_t = a_i$, the policy gradient is
\begin{equation}
\frac{\partial L}{\partial z_j} = \hat{A}_t \, (\mathbf{1}_{i=j} - \pi_j),
\end{equation}
where $\mathbf{1}_{i=j}$ is the indicator function. Suppose an action $a_k$ is invalid. The gradient for this invalid action is then:
\begin{equation}
\frac{\partial L}{\partial z_k} = \hat{A}_t (0 - \pi_k) \approx 0,
\end{equation}
So, no gradient flows in the case of an invalid action. In contrast, for a valid action $a_i$ that was sampled and resulted in an negative advantage $\hat{A}_t < 0$, the gradient pushes the logit $z_i$ down, decreasing its probability in the future. Consequently, valid actions with negative advantage are punished more than invalid actions. 

\begin{equation}
\frac{\partial L}{\partial z_i} = \hat{A}_t (1 - \pi_i) < 0
\end{equation}

\subsection{Gradient-based action masking}

We propose a gradient-based approach that directly modifies the policy network by assigning a negative gradient to invalid actions, rather than relying solely on an external mask. This encourages the agent to suppress invalid moves during training and enables the policy to implicitly learn the feasibility constraints encoded by the Petri net guard functions.

Let $\pi_\theta(a|s)$ denote the policy probability for action $a$ in state $s$, parameterized by $\theta$, and let $\mathcal{A}_\text{inv}(s)$ be the set of invalid actions according to the Petri net guard function. After computing the standard policy gradient loss $L_\text{PG}$ (e.g., from PPO), we define an additional invalid-action penalty:

\begin{equation}
L_\text{inv}(\theta) = \lambda_\text{inv} \sum_{a \in \mathcal{A}_\text{inv}(s)} \pi_\theta(a|s),
\end{equation}

where $\lambda_\text{inv} > 0$ is a tunable hyperparameter controlling the strength of the penalty. The total policy loss becomes:

\begin{equation}
L_\text{total}(\theta) = L_\text{PG}(\theta) + L_\text{inv}(\theta).
\end{equation}

The corresponding gradient update naturally pushes the logits of invalid actions downward:

\begin{equation}
\nabla_\theta L_\text{inv}(\theta) = \lambda_\text{inv} \sum_{a \in \mathcal{A}_\text{inv}(s)} \nabla_\theta \pi_\theta(a|s) < 0.
\end{equation}

By consistently penalizing invalid actions during training, the policy learns to approximate the guard function of the Petri net: in states where an action is invalid according to the guard, the network assigns near-zero probability, effectively internalizing the feasibility constraints. Over time, this reduces the reliance on an external mask, allowing the agent to respect structural constraints autonomously while improving sample efficiency and stability in highly constrained environments like dynamic JSSPs.

\section{Dynamic Fault Tolerant PetriRL}
\label{sec: dynamic_fault_tolerant}

In this chapter, we present our methodology for handling dynamic events in the JSSP using a CPTN  combined with a maskable Actor-Critic RL agent. We start by providing an overview of the interaction between the main framework elements, particularly the environment and the RL agent. Then we delve into detailing every component in the subsequent chapters.

\begin{figure*}[ht]
\centering
\includegraphics[width=\linewidth]{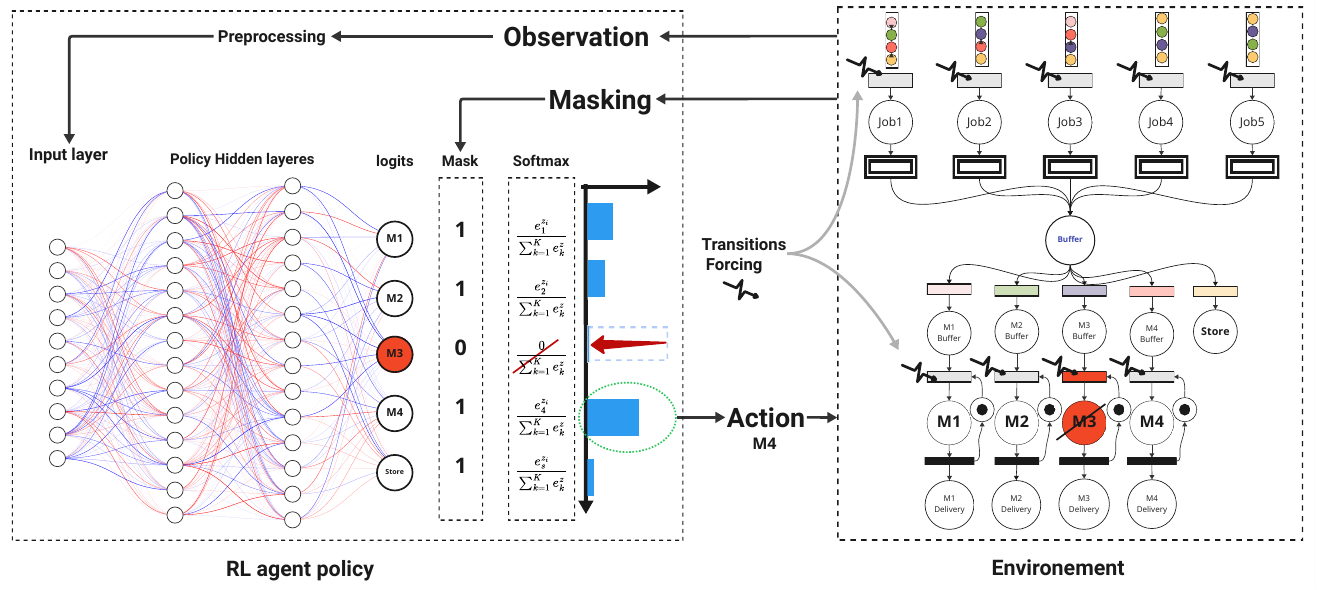}
\caption{Illustration of the synergistic interaction between Petri net guard functions and the action masking mechanism in the policy network.}
\label{fig:frameworkt}
\end{figure*}

Figure~\ref{fig:frameworkt} provides an overview of the proposed framework. The agent interacts with the Petri net through controllable transitions, with valid actions determined by the Petri Net's guard functions and external constraints. The environment provides a Boolean mask that filters out invalid transitions, allowing the agent to focus only on feasible decisions. This reduces the adequate action space, improves credit assignment, and enhances training efficiency.

In this paper, we explore manual enforcement of transition states alongside automatic token-based masking. Forced transition states enable the modeling of dynamic scenarios such as machine breakdowns and random job arrivals. For instance, when a machine is disabled, its associated probability in the RL policy network is reduced to zero, and the agent automatically selects the next best feasible action based on the current state. This mechanism underlies the concept of fault tolerance: regardless of how many machines are disabled, the agent will always choose the next best feasible action.

Furthermore, unlike many fault-tolerant strategies in the literature, which assume non-resumable jobs, our simulation models machine breakdowns with resumable operations. When a machine is down, it cannot accept new operations, and any token currently in progress is paused. Once the machine returns to operation, processing resumes from where it was halted, better reflecting real-world behavior.

Two distinct masking strategies can be distinguished during agent training in the Petri net environment: a gradient-based and a gradient-free approach. In the first gradient-free approach, manual masking simulates machine breakdowns. As an external event rather than a system constraint,  learning the masking behavior offers no benefit to the agent and serves solely as a simulation tool.. In the second case, masking arises naturally from the Petri net’s guard functions. This masking reflects genuine environmental constraints and the scheduling rules. Learning to respect these masks can be advantageous for the agent, enabling it to avoid invalid actions without external enforcement.

After providing an overview of our approach, the next chapter delves into the details: modeling the JSSP with CTPN, simulating random events such as machine breakdowns and job arrivals, and finally, the choice of RL algorithm along with its key elements such as observation, action space, and reward function.

\subsection{Environment using Colored-Timed Petri Nets}
\label{subsec: Environment}

\begin{figure}[ht]
\centering
\includegraphics[width=0.8\linewidth]{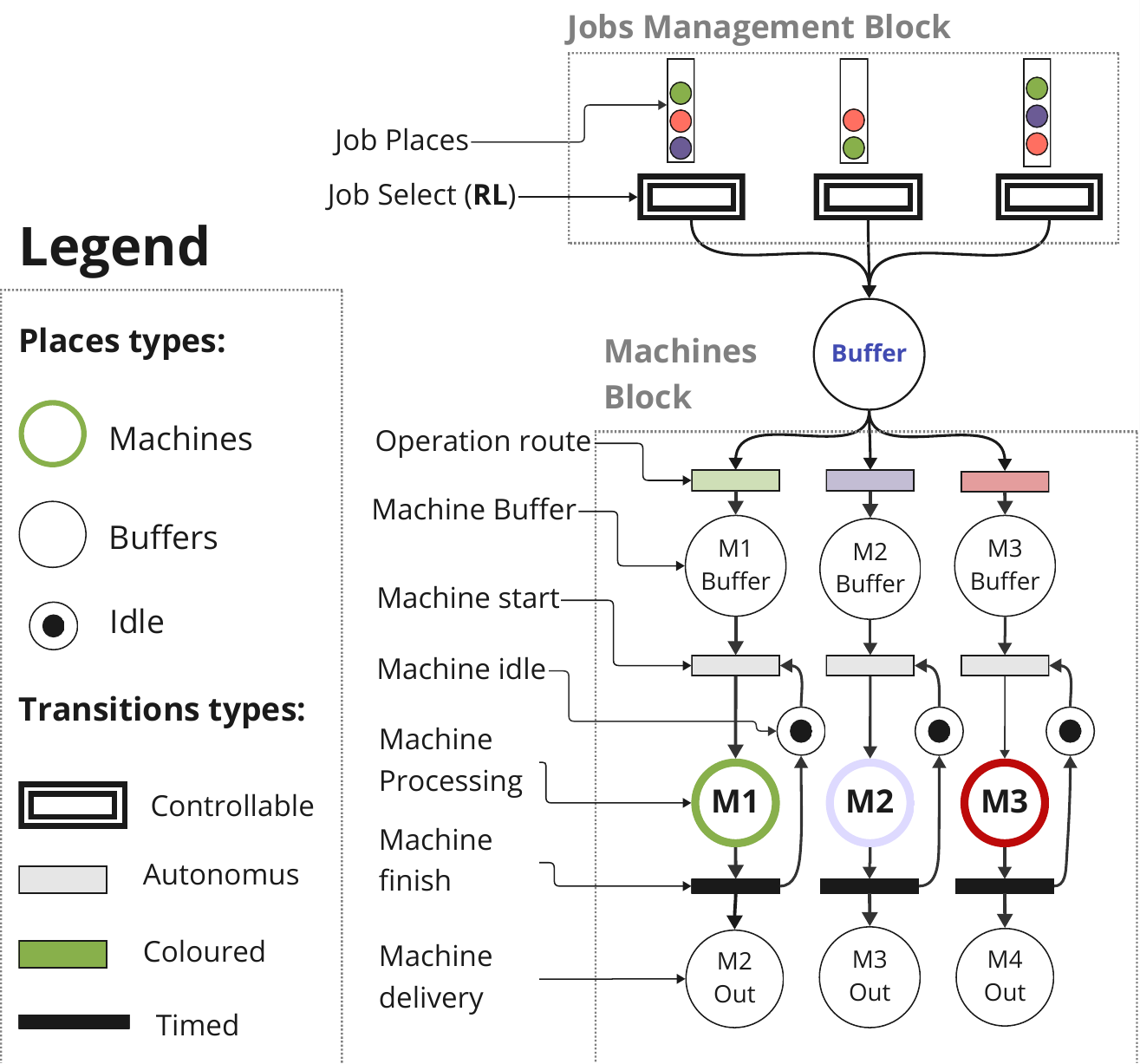}
\caption{Colored Timed Petri Net modeling a Job Shop Scheduling Problem with 3 jobs and 3 machines.}
\label{fig: dynamic JSSP model}
\end{figure}

In the RL framework, the agent continuously interacts with the environment to improve its policy. We model the environment using CTPN  depicted in Fig.\ref{fig: dynamic JSSP model}, which offers formal structure and interpretability. By combining Petri nets with the adaptive capabilities of RL, we develop an effective solver for the DJSSP. 

The Petri net paradigm, introduced in Section \ref{subsec: colored-Timed Petrinets}, is built upon three fundamental components: places, transitions, and tokens. The flow and distribution of tokens primarily govern the system dynamics. In the context of the JSSP, each job comprises a predefined sequence of operations subject to precedence constraints. In our model, these operations are represented as a series of colored tokens, where each token encodes machine compatibility through its color as well as key operational attributes such as processing time.

The model incorporates multiple places with different roles to represent the state of the system. All places follow a First-In-First-Out policy, ensuring that tokens are consumed in the order they arrive. These include job places, which hold operation tokens; machine places, which reflect machine availability and utilization; and additional organizational places, such as idle machine places and buffer places, which support the coordination and flow of jobs throughout the system.

In addition to tokens and places, we define four transition types: autonomous, controllable, colored, and timed. A transition fires when its specific conditions are met, typically when all input places hold at least one token, along with any additional constraints based on its type. Controllable transitions, such as job selection, require an additional external trigger, in our case, a decision from the RL agent. Colored transitions, mainly used for routing, fire only when the transition's color matches that of the token. Timed transitions fire automatically once the token has spent the required sojourn time in the corresponding processing place.

The token flow in our model can be described as follows:  The RL agent determines the sequence of operations by firing controllable "job selection" transitions, which move operation tokens from their job places to a routing buffer. There, colored "operation route" transitions direct each token to the appropriate machine buffer based on its color. If the machine is idle and a token is present in the buffer, the autonomous "machine start" transition fires, moving the token to the machine processing place. After the token spends its required sojourn time, the timed "machine finish" transition fires, transferring the token to the machine delivery place.

\subsection{Modeling Machine Breakdowns Using the Weibull Distribution}
\label{subsec: Incorporation of scheduled machine downtimes}

After modeling the job shop floor with CTPN, we simulate machine breakdowns and random job arrival events. Starting with the machine breakdown events, we choose to model them using the Weibull distribution, as it better captures aging and wear-out effects than exponential or Poisson models \cite{McCool.2012}. The probability density function of the Weibull distribution is given by:

\begin{equation*}
\tag{7}
f(t) = \frac{\beta}{\eta} \left( \frac{t}{\eta} \right)^{\beta - 1} e^{-(t/\eta)^\beta}, \quad t \geq 0
\end{equation*}
where:\\
\noindent
-- \(\beta > 0\) is the \textit{shape parameter},\\
-- \(\eta > 0\) is the \textit{scale parameter},\\
-- \(t\) is the time to failure.
\vspace{1em}

\begin{figure}[ht]
\centering
\includegraphics[width=1\linewidth]{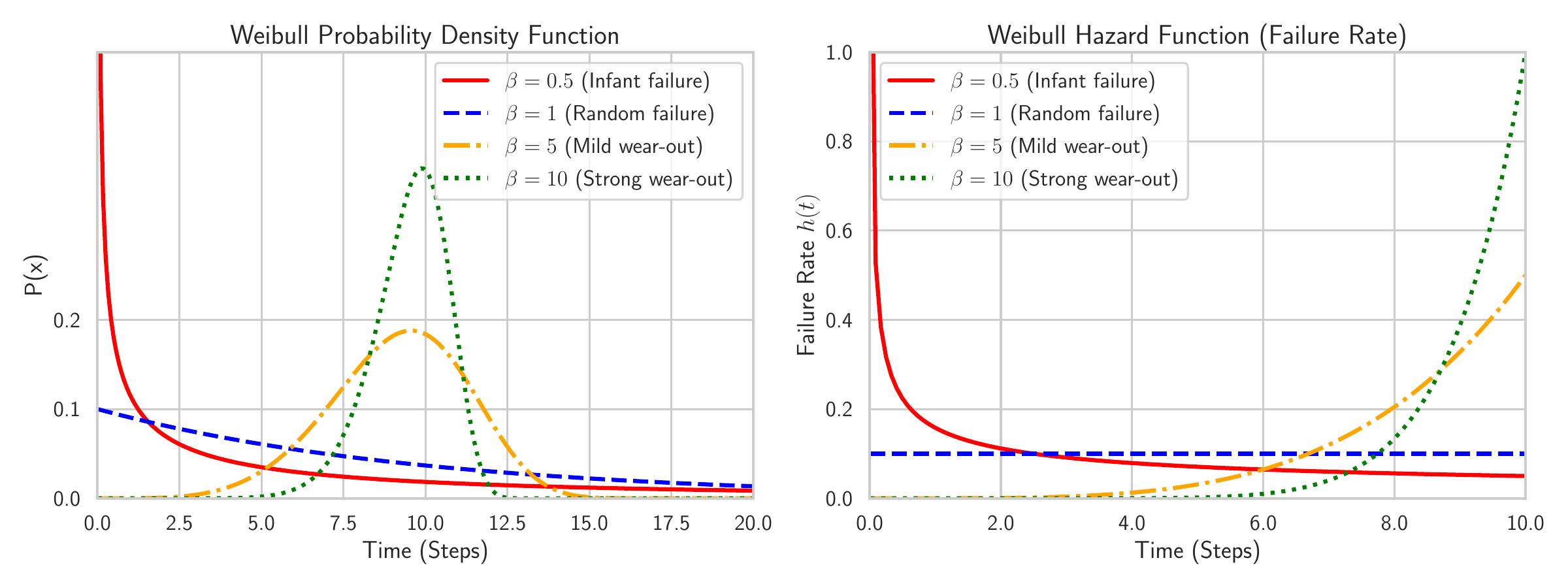}
\caption{Top: Probability density functions of the Weibull distribution for different shape parameters \(\beta\). The shape \(\beta = 0.5\) models infant mortality \(\beta = 1\) models random failure with a constant rate, and \(\beta = 5\) models wear-out failure. Bottom: The corresponding hazard functions \(h(t)\), which illustrate how the failure rate evolves depending on the value of \(\beta\).}
\label{weibull distribution}
\end{figure}

The Weibull distribution is widely used in reliability applications due to its flexible shape parameter \(\beta\). For \(\beta < 1\), it models infant mortality with a decreasing failure rate, useful for new equipment. For \(\beta > 1\), it captures aging with an increasing failure rate, where higher \(\beta\) indicates more aggressive aging. When \(\beta = 1\), it simplifies to the exponential distribution with a constant failure rate. This behavior is illustrated in Fig.~\ref{weibull distribution}, especially in the hazard function. We also used the scale parameter $\eta$ to ensure that breakdown events have a high probability during the processing time of a job operation given by the instance.

Now that we have a mathematical model for machine breakdowns, we proceed to simulate these events within our system. Machine unavailability is implemented by forcing the machine's start transition to be disabled. The failure time step is sampled from a Weibull distribution, while the repair duration follows a Normal distribution. 

If a breakdown occurs while a token is being processed, the processing timer is paused and resumes only once the machine is restored. This interruption affects both routing and execution: transitions to the failed machine are blocked, and job progress is stalled due to precedence constraints; no operation may begin until its predecessor completes. If a job is held on a failed machine, its entire sequence is delayed. Notably, this behavior emerges naturally from the Petri net's token dynamics, requiring no explicit enforcement of additional constraints.

%In practice, modeling the failure probability begins with data collection, followed by either graphical or analytical methods to fit the Weibull distribution and estimate the parameters. Analytical approaches include the maximum likelihood method, least squares, and the method of moments. The authors in \cite{Hatakeyama.2014}  presented complete parameter estimation procedures. In this work, we assume the parameters are known and inject them directly into our simulation. Based on these parameters, we compute the machine failure times and recovery times to determine when each machine becomes available again. In the next chapter, the details of the modeling of the random job arrival events will be discussed.

\subsection{Modelling the random job arrivals }
\label{subsec: Modelling the random job arrivals}

In many production systems, job arrivals are often modeled using a Poisson process due to its simplicity. However, this approach assumes that the inter-arrival times between job operations follow a constant arrival rate. Such assumptions are often too restrictive for real-world systems, where arrival patterns may exhibit burstiness, clustering, or irregular delays.

\begin{figure}[ht]
\centering
\includegraphics[width=0.9\linewidth]{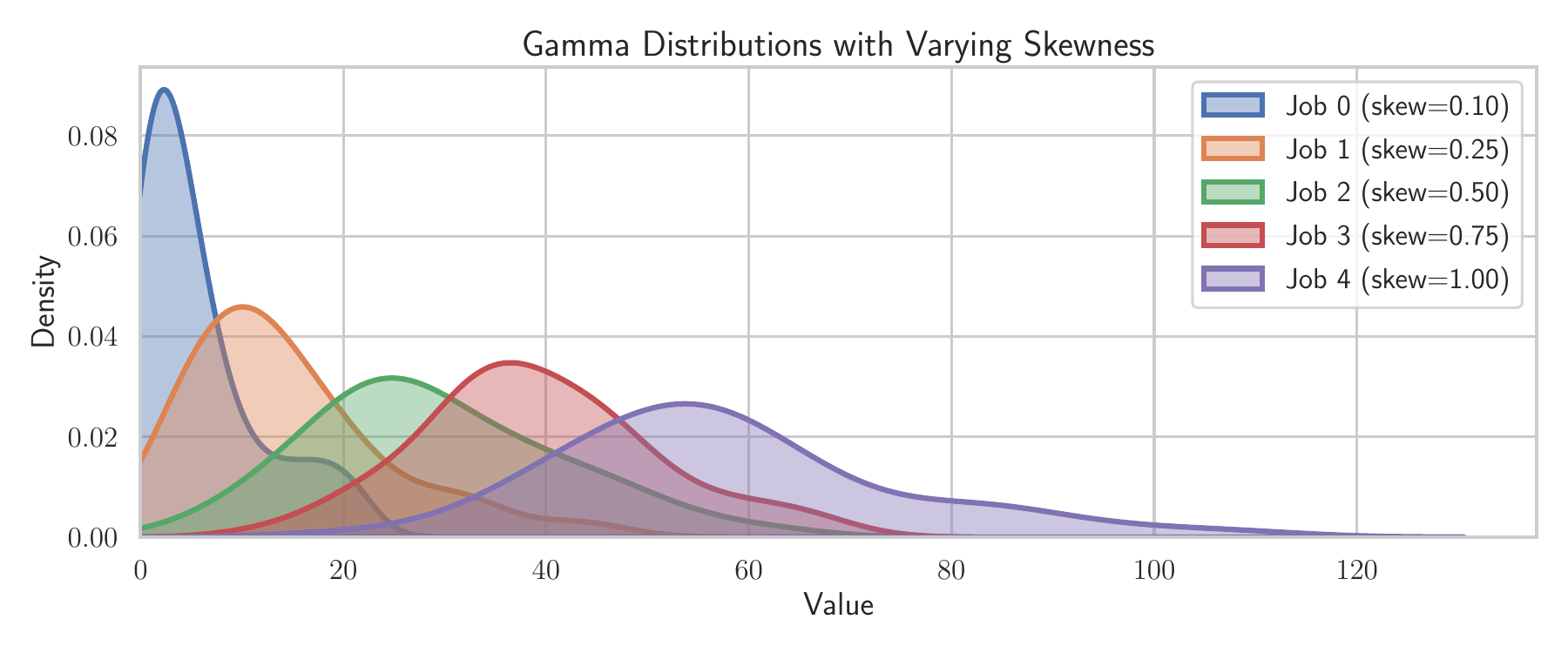}
\caption{Gamma-distributed inter-arrival times for five jobs with increasing skewness. Higher skewness captures bursty or irregular job arrivals, while lower skewness reflects more regular patterns.}
\label{fig: gamma destribution}
\end{figure}

To better capture this variability, we model the inter-arrival times using the Gamma distribution, which generalizes the exponential distribution by introducing a shape parameter \(\alpha\) and a scale parameter \(\beta\). The Gamma distribution provides additional flexibility by allowing us to control both the mean and variance of arrival intervals, enabling more realistic modeling of job dynamics. Let \(T_i\) denote the time between the arrivals of successive operations. We assume:

\[
T_i \sim \text{Gamma}(\alpha, \beta),
\]

\noindent where \(\alpha > 0\) is the shape parameter controlling the skewness of the distribution, and \(\beta > 0\) is the scale parameter. The corresponding probability density function is given by:

\begin{equation}
f(t; \alpha, \beta) = 
\frac{t^{\alpha - 1} e^{-t/\beta}}{\beta^{\alpha} \Gamma(\alpha)}, \quad t \geq 0.
\tag{8}
\end{equation}

In our implementation, each job is assigned a random skewness value \( s \in [0,1] \), which parameterizes the gamma distribution with shape \(\alpha = 10s\) and scale \(\beta = 0.1H\), where \(H\) is the planning horizon. The choice of \(\beta = 0.1H\) ensures that the majority of arrivals occur at the beginning of the job planning horizons. The shape parameter \(\alpha\) controls the pattern of job arrivals, with higher \(s\) yielding more tightly clustered operations. In Fig.\ref{fig: gamma destribution}, we illustrate an example of various gamma distributions modeling five random job arrival distributions with different behaviors. 

Random job arrivals are modeled using a dedicated planned jobs place. Unlike the static case in the traditional JSSP, where all jobs are available upfront, operations are released dynamically via a job release transition, which transfers tokens from the planned jobs place to the ready-for-selection place. This transition is forced to fire based on a Gamma distribution associated with a given job. At each time step, the distribution is sampled to determine how many tokens are released.

\subsection{RL agent key elements and parameters}

Now that the environment has been defined, including the modeling of random job arrivals and machine downtimes, we shift our focus to the key design choices and parameters of the RL agent. We begin by selecting the appropriate RL algorithm for the agent, followed by a detailed description of the action space, observation space, and reward function.

For the RL agent, we adopt the Maskable variant of Proximal Policy Optimization (PPO). This choice is motivated by two key reasons. First, PPO is a robust and stable algorithm, owing to its actor-critic architecture and the use of a clipping function that prevents large, destabilizing policy updates during training. Second, and most critically for our setting, the actor network in PPO can easily incorporate dynamic action masking. This capability aligns well with our approach, where the Petri net dynamically provides a list of valid actions at each time step. By masking out invalid actions, the RL agent significantly reduces the effective action space, an essential advantage in highly constrained problems such as the JSSP. This approach not only improves credit assignment but also increases training efficiency by concentrating policy updates on feasible and relevant decisions. The implemented algorithm is detailed in Algorithm~\ref{algorithm: maskable PPO}.

\begin{algorithm}[H]
\caption{Proximal Policy Optimization with Action Masking via Petri Net Guards}
\label{algorithm: maskable PPO}
\begin{algorithmic}[1]
\State \textbf{Input:} Initial policy parameters $\theta$, value function parameters $\phi$, clipping parameter $\epsilon$, learning rate $\alpha$, balancing coefficient $C_1$
\For{iteration $= 1,2,\ldots$}
    \State Collect trajectories $\{(s_t, a_t, r_t, s_{t+1})\}$ using $\pi_\theta$
    \State Compute advantages $\hat{A}_t$ and returns $R_t$
    
    \State Compute value function loss: 
    $L^{VF}(\phi) = \mathbb{E}_t \big[ (V_\phi(s_t) - R_t)^2 \big]$
    \State Update $\phi$ via gradient descent
    
    \State Compute action mask: $\text{Mask}[i] = G(a_i), \forall a_i \in \mathcal{T}_a$
    
    \State Compute masked policy logits:
    $z_t(a) = \begin{cases}
        \text{policy\_network}(s_t, a), & G(a) = 1\\
        -\infty, & \text{otherwise}
    \end{cases}$

    \State Compute clipped surrogate objective: 
    $L^{CLIP}(\theta) = \mathbb{E}_t \big[ \min(\rho_t \hat{A}_t, \text{clip}(\rho_t,1-\epsilon,1+\epsilon)\hat{A}_t) \big]$
    
    \State Total PPO loss: $L(\theta) = L^{CLIP}(\theta) + C_1 L^{VF}(\phi)$
    \State Update policy: $\theta \leftarrow \theta - \alpha \nabla_\theta L(\theta)$
\EndFor
\end{algorithmic}
\end{algorithm}

The observation and action space are directly derived from the Petri net model. The observation corresponds to the current marking, which reflects the token distribution across places and captures the system state. We augment this with additional attributes such as token colors and remaining processing times. While observations are based on places, actions correspond to transitions. At each step, the Petri net uses guard functions and token distributions to compute the set of valid transitions, dynamically defining the action space. The RL agent selects one valid transition to fire, updating the system state and repeating the cycle. An episode terminates when all tokens from the planned job places reach their delivery places and all intermediate places are empty.

Many objectives can be considered in scheduling optimization, such as minimizing makespan, meeting due dates, or maximizing machine utilization. In this work, we focus primarily on minimizing the \textbf{makespan} as the main evaluation metric. To guide the agent toward this objective, we define the reward function so that the agent is penalized by the final makespan of the schedule.

Formally, the reward at time step \( t \) is defined as:

\begin{equation}
R_t = 
\begin{cases}
0, & \text{if } t < T \\
-\text{Makespan}, & \text{if } t = T
\end{cases}
\tag{9}
\end{equation}

where \(T\) is the terminal time step corresponding to the completion of the scheduling process. Penalizing the makespan yields sparse rewards and naturally leads to the credit assignment problem; yet, in our experiments, this approach consistently outperformed alternatives that provide more frequent feedback, such as per-step machine utilization. This is because reward shaping carries the risk of reward hacking, in which the agent optimizes for intermediate criteria rather than improving the overall objective.

\section{Results and Discussion}
\label{sec: Results and Discussion}

This section presents our experimental results. We introduce the baseline methods, benchmark instances, and evaluation metrics. After detailing the experimental setup, we analyze the training performance of the RL agent, discuss the results, and conclude with an ablation study on key components of our approach.
\subsection{Benchmarks and evaluation metrics}
\label{subsec: Benchmarks}

We benchmarked our approach against 14 dispatching heuristics presented in \cite{ArickoKhenaKaban.2012}, which are categorized into static and dynamic rules. Static rules, such as SPS (Shortest Processing Sequence), prioritize jobs with the shortest sequences, so it bases its decision on the initial state of a job. In contrast, dynamic rules adapt during the simulation based on the current system state, for instance, MTWR (Most Total Work Remaining), which selects jobs with the highest cumulative remaining processing time. 

The decision to benchmark primarily against dispatching rules is driven by two key factors: the dynamic nature of the problem and reproducibility challenges. First, the dynamic environment requires real-time solution updates in response to changing conditions, making exact solvers and metaheuristics impractical. This limits viable approaches to rule-based and policy-based methods. Second, reproducibility is a concern, as most existing studies evaluate performance on fixed, controlled scenarios (e.g., Taillard instances). To fairly compare results, full access to and control over other simulators are needed to replicate the same random conditions. One of the goals of this paper is to address this gap by releasing an open-source, Gym-compatible environment to support reproducible research.

To ensure a fair comparison between heuristic rules and the RL agent, each acting as a decision-maker, they interact with the same Petrinet-based environment. At every interaction step, a list of valid actions is dynamically filtered using the Petri net's guard functions. This list is then provided to the decision-making module. The RL agent uses its policy network to select an action from this list, while the heuristic methods apply their predefined dispatching rules to make their choice. The selected action is subsequently executed in the environment. Once the episode is over and all the jobs are processed, the makespan is used as the main evaluation metric across the different optimizers. The list of employed heuristics is provided in Table~\ref{tab:heuristics}.   

\begin{table}[ht]
  \centering
    \begin{tabular}{lc}
      \toprule
      \multicolumn{2}{c}{\textbf{Job Selection Heuristics}} \\
      \midrule
      FIFO  & Job entered the system earliest \\
      SPS   & Job with shortest processing sequence \\
      LPS   & Job with longest processing sequence \\
      SPTN  & Job with shortest next operation time \\
      LPTN  & Job with longest next operation time \\
      MTWR  & Job with most total work remaining \\
      LTWR  & Job with least total work remaining \\
      LWT   & Job with longest waiting time \\
      SPT   & Job with shortest total processing time \\
      LPT   & Job with longest total processing time \\
      SPSR  & Job with shortest remaining processing sequence \\
      LPSR  & Job with longest remaining processing sequence \\
      \bottomrule
    \end{tabular}%
  \caption{List of heuristics used for benchmarking.}
  \label{tab:heuristics}
\end{table}

Modeling stochastic elements such as job arrivals and machine breakdowns while maintaining fair benchmarking across optimization approaches is a non-trivial task. To address this, we adopt two complementary strategies. First, each experimental configuration is executed over 100 independent runs, and we report the average performance along with statistical measures such as mean, variance, and 95\% confidence intervals. 

Second, to guarantee reproducibility and fairness across algorithms, we use controlled randomness via a predefined list of 100 unique seeds, one per run. These seeds are used to initialize the random number generators responsible for sampling stochastic events, such as job arrival times and machine breakdowns, which follow Weibull or Gamma distributions detailed in Section \ref{sec: problem_formulation}. Crucially, the same sequence of seeds is used for all optimization methods. For instance, the first run of any tested algorithm always uses the first seed, the second run the second seed, and so on. This approach ensures that each algorithm is exposed to the same scenario, allowing for consistent, reproducible, and fair comparisons across different solution methods.

\subsection{Experimental Setup}
\label{subsec: Experimental setup}

\begin{figure}[ht]
\centering
\includegraphics[width=0.7\linewidth]{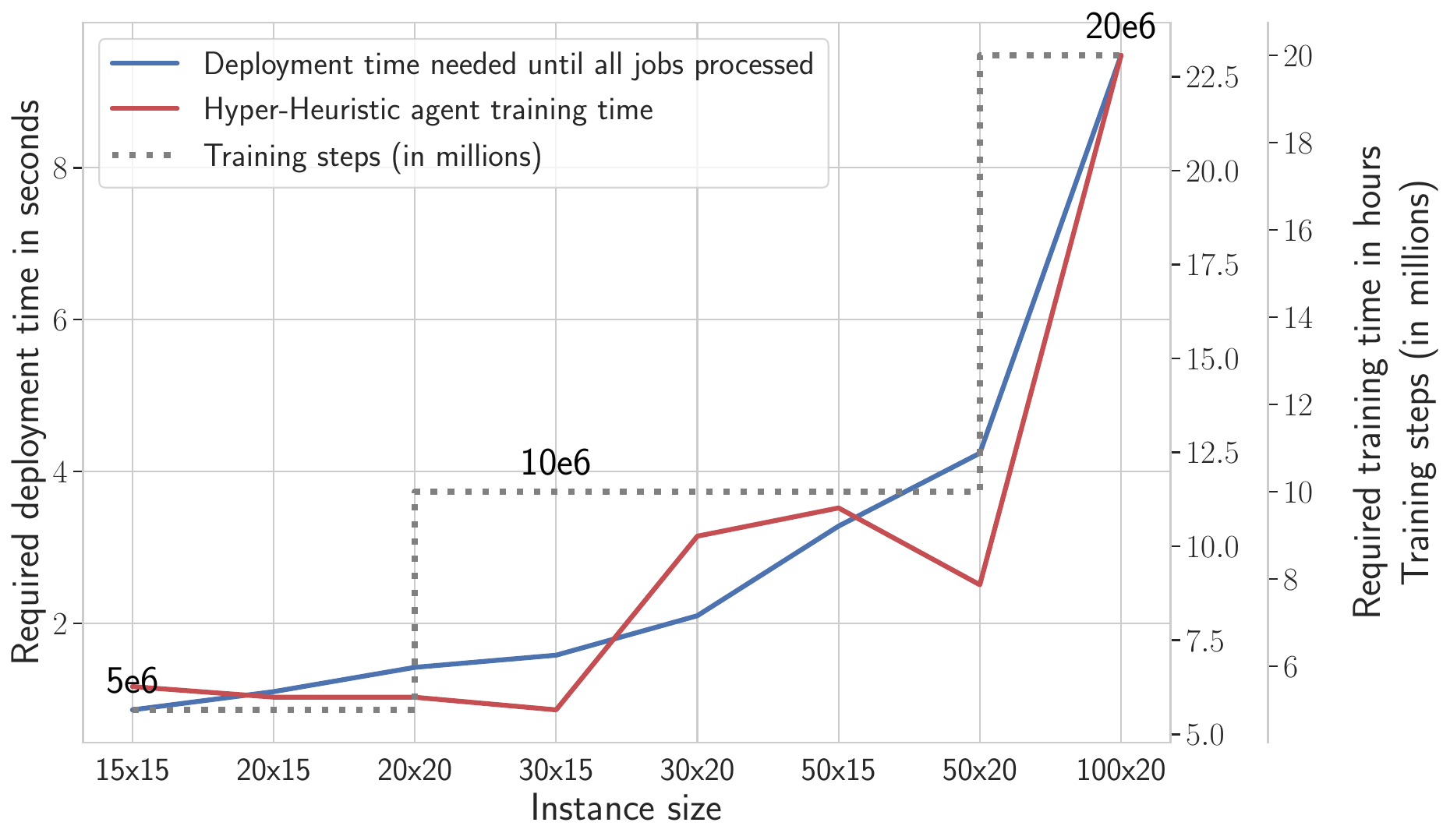}
\caption{Training and deployment time for different instances.
The red line represents the agent training time requirement for different instance sizes, as shown on the corresponding y-axis on the right.
The blue line represents the time required to solve the DJSSP problem during the inference. The corresponding y-axis is on the left.
The dotted grey line represents the increasing number of training steps in every size group.}
\label{fig: time_log}
\end{figure}

All experiments were conducted on a system running Windows 11, equipped with an NVIDIA Quadro RTX 5000 GPU. The models were implemented using the PyTorch deep learning framework in combination with the Stable-Baselines3 library \cite{Hill.2018}. Details regarding the number of training steps, total training time, and deployment time are presented in Figure \ref{fig: time_log}

We assessed performance across different problem scales using two benchmark sets. The first is a reduced-scale set based on the Raj benchmark \cite{Raj.2014}, consisting of instances with up to 10 jobs and 5 machines. The second is the widely used Taillard benchmark \cite{Eric.1993}, which includes 80 instances grouped into seven size categories, ranging from 15 jobs and 15 machines to 100 jobs and 20 machines. Each group contains 10 instances of the same size, populated with different processing times and job sequences.

\subsection{RL Agents Training}
\label{subsec: RL Agents Training}

\begin{figure}[H]
\centering
\includegraphics[width=\linewidth]{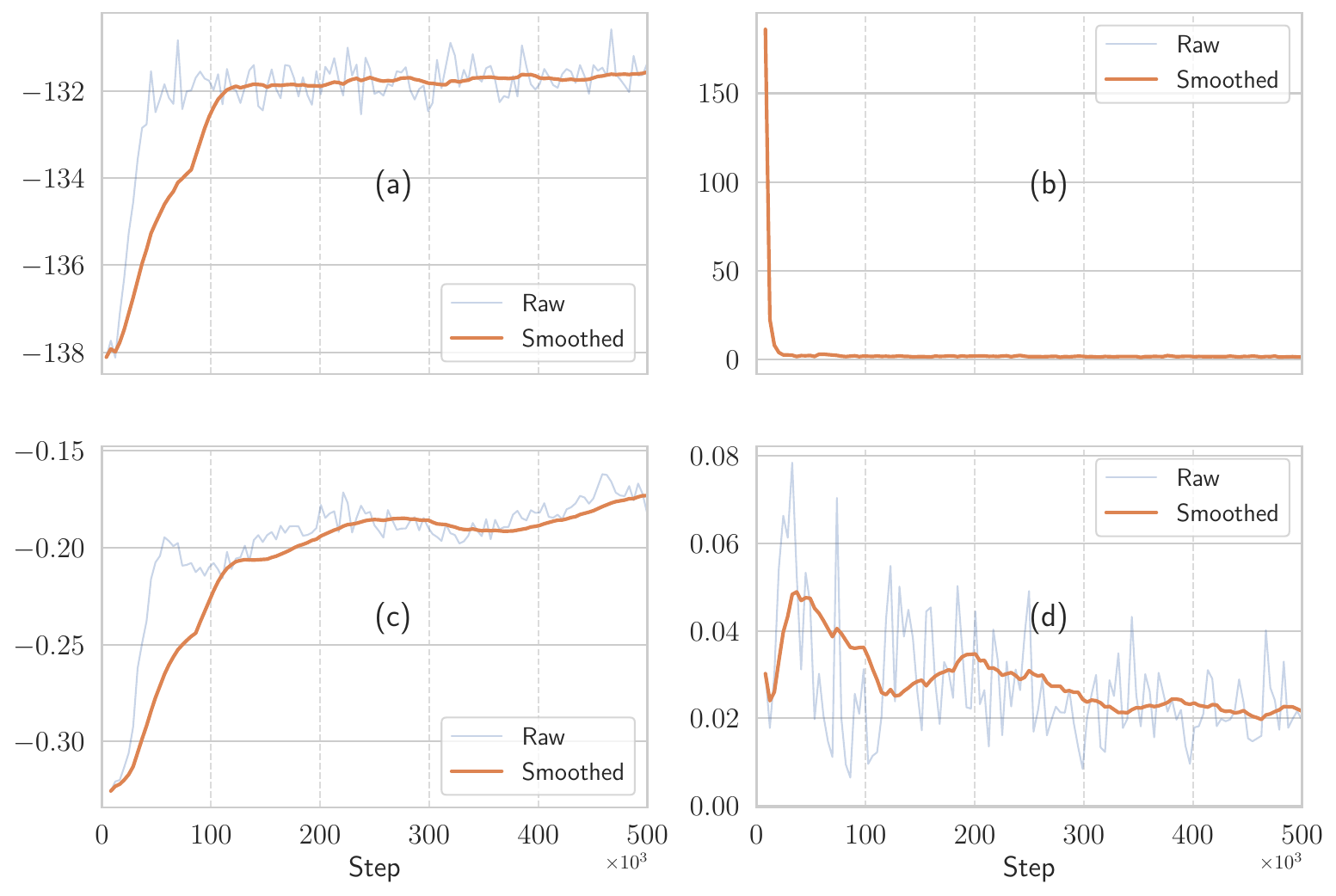}
\caption{Agent training performances on a 5 jobs x 4 machines instance . (\textbf{a})  the episode mean reward  , (\textbf{b}) the combined training loss,(\textbf{c}) the entropy loss (\textbf{d}) the clipping range. The agent is trained for \textbf{5e5} steps using the maskable PPO algorithm.}
\label{fig: training_performances}
\end{figure}

Fig.~\ref{fig: training_performances} shows the training performance of an agent trained on a 5-jobs × 4-machines instance. We selected four key metrics to evaluate training: the episode mean reward, combined training loss, entropy loss, and clipping range.

The episode mean reward shown in subplots (a) steadily increases and stabilizes around a makespan of 132 without any intermediate plateaus, indicating that the reward function effectively guides the agent to improve its results. In PPO, the combined loss shown in subplots (b) aggregates the policy loss, value loss, and entropy bonus. The value loss reflects how well the critic network predicts state values, while the policy loss shows that the new policy improves action probabilities when associated with positive advantages. Together, the reward and loss curves provide insight into overall training progress.

The entropy loss and clipping range, shown in subplots (c) and (d), respectively, reflect the exploration-exploitation tradeoff. The entropy loss decreases steadily from a high absolute value, indicating that the agent transitions from initial exploration to exploitation by adopting a more deterministic policy. Meanwhile, the clipping range decreases, showing that the agent makes smaller policy updates as training progresses, further signifying a shift toward more stable and confident behavior.

Overall, the training metrics suggest that the agent learns a stable and effective policy, balancing exploration and exploitation successfully.

\subsection{Results under Machine breakdown }
\label{subsec: Results under Machine breakdown }

\begin{table*}[ht]
\centering
\scriptsize
\setlength{\tabcolsep}{3.5pt}
\resizebox{\textwidth}{!}{%
\begin{tabular}{@{}lcccccccccccccccc@{}}
\toprule
Instance & FIFO & SPT & LPT & SPS & LPS & LTWR & MTWR & SPSR & LPSR & SPTN & LPTN & LWT & \makecell{Heur.\\Avg.} & \makecell{Best\\Heur.} & Ours & Gap \% \\ 
\midrule
ra01 & 69.1 & 69.2 & 74.8 & 69.2 & 69.1 & 69.2 & 74.8 & 69.2 & 69.1 & 69.2 & 74.8 & 69.1 & 70.6 & 69.1 & \textbf{69.1} & 2.1 \\
ra02 & 81.9 & 81.9 & 88.1 & 81.9 & 88.2 & 81.9 & 88.1 & 81.9 & 88.2 & 81.9 & 81.9 & 81.9 & 84.0 & 81.9 & \textbf{78.7} & 6.2 \\
ra03 & 89.6 & 96.0 & 84.9 & 89.6 & 84.9 & 96.0 & 84.9 & 89.6 & 84.9 & 89.1 & 89.6 & 89.6 & 89.1 & 84.9 & \textbf{84.9 }& 4.7 \\
ra04 & 63.1 & 65.2 & 63.0 & 63.1 & 63.0 & 65.2 & 64.0 & 63.1 & 63.0 & 63.0 & 64.0 & 63.1 & 63.6 & 63.0 & \textbf{63.0} & 0.9\\
ra05 & 55.9 & 55.9 & 59.8 & 55.9 & 55.9 & 55.9 & 59.8 & 55.9 & 55.9 & 55.9 & 59.8 & 55.9 & 56.8 & 55.9 & \textbf{55.9} & 1.7 \\
ra06 & 103.3 & 104.1 & 103.7 & 103.3 & 103.3 & 104.1 & 103.7 & 103.3 & 103.3 & 104.1 & 103.7 & 103.3 & 103.6 & 103.3 &\textbf{ 98.5} & 4.9\\
ra07 & 74.8 & 74.8 & 73.8 & 74.8 & 75.7 & 74.8 & 73.8 & 74.8 & 75.7 & 74.8 & 74.1 & 74.8 & 74.7 & 73.8 &\textbf{ 73.8} & 1.3 \\
ra08 & 150.5 & 150.5 & 150.5 & 150.5 & 150.5 & 150.5 & 150.5 & 150.5 & 150.5 & 150.5 & 150.5 & 150.5 & 150.5 & 150.5 & \textbf{150.5} & 0.0 \\
ra09 & 95.6 & 95.6 & 98.1 & 93.4 & 97.9 & 95.6 & 98.1 & 93.4 & 97.9 & 97.9 & 96.1 & 95.6 & 96.3 & 93.4 & \textbf{93.4} & 3.0\\
ra10 & 137.2 & 141.2 & 135.3 & 135.4 & 135.3 & 141.2 & 135.3 & 135.4 & 135.3 & 137.5 & 136.9 & 137.2 & 137.0 & 135.3 & \textbf{131.2 }& 4.2 \\
ra11 & 82.3 & 92.3 & 82.3 & 87.4 & 82.3 & 92.3 & 82.3 & 88.0 & 82.3 & 92.3 & 82.3 & 82.3 & 85.7 & \textbf{82.3} & 82.7 & 3.5 \\
ra12 & 70.8 & 70.8 & 70.2 & 70.8 & 70.2 & 70.8 & 70.2 & 70.8 & 70.2 & 70.4 & 71.4 & 70.8 & 70.6 & 70.2 & \textbf{70.1} & 0.7 \\
ra13 & 121.1 & 123.2 & 121.5 & 121.1 & 127.1 & 123.2 & 121.5 & 121.1 & 127.1 & 123.6 & 131.0 & 121.1 & 123.5 & 121.1 & \textbf{121.1} & 2.0 \\
ra14 & 76.3 & 76.3 & 66.0 & 76.3 & 66.0 & 76.3 & 66.0 & 76.3 & 66.0 & 73.2 & 76.3 & 76.3 & 72.6 & 66.0 & \textbf{63.3} & 12.8 \\
ra15 & 82.7 & 82.7 & 92.3 & 92.7 & 82.7 & 82.7 & 91.5 & 92.7 & 82.7 & 82.7 & 92.7 & 82.7 & 86.7 & 82.7 & \textbf{82.7 }& 4.6 \\
ra16 & 77.5 & 77.4 & 81.9 & 77.5 & 81.8 & 77.4 & 81.9 & 77.5 & 81.8 & 78.4 & 79.4 & 77.5 & 79.2 & 77.4 & \textbf{77.2 }& 2.4 \\
ra17 & 60.8 & 66.8 & 65.1 & 66.8 & 60.8 & 66.8 & 65.1 & 66.8 & 60.8 & 66.8 & 65.8 & 60.8 & 64.5 & 60.8 & \textbf{60.8 }& 5.7 \\
ra18 & 69.7 & 69.7 & 69.7 & 69.7 & 69.7 & 69.7 & 69.7 & 69.7 & 69.7 & 69.7 & 69.7 & 69.7 & 69.7 & 69.7 & \textbf{69.7} & 0.0 \\
ra19 & 94.1 & 93.3 & 92.1 & 94.1 & 91.0 & 93.3 & 92.1 & 94.1 & 90.6 & 90.6 & 94.1 & 94.1 & 92.8 & 90.6 & \textbf{90.5} & 2.4 \\
ra20 & 79.0 & 79.0 & 79.2 & 79.0 & 79.3 & 79.0 & 79.2 & 79.0 & 79.3 & 79.0 & 79.2 & 79.0 & 79.1 & 79.0 & \textbf{79.0} & 0.1\\
\midrule
Avg. & 86.8 & 88.3 & 87.6 & 87.6 & 86.7 & 88.3 & 87.6 & 87.7 & 86.7 & 87.5 & 88.7 & 86.8 & 87.5 & 85.5 & \textbf{84.8} & 3.0 \\
\bottomrule
\end{tabular}}
\caption{Performance comparison between heuristic methods and our approach on small-sized instances from the Raj benchmark under machine breakdown scenarios.}
\label{tab: raj machine only}
\end{table*}

We start our results discussion with the Machine breakdown scenarios. In Table \ref{tab: raj machine only}, we present the results of our approach compared to twelve heuristic methods under random machine breakdown scenarios for the Raj instances. As described in the experimental setup, all algorithms experience the same exact events. To further minimize the impact of randomness, we sampled 100 runs for each configuration and reported the mean values. We also report an average 95\% confidence interval of ±0.9 time steps across all runs.

The results show that our approach consistently outperforms all heuristics. We compared our results both to the average performance of all heuristics and to the best-performing heuristic for each instance. Our approach achieved results equal to or better than the best heuristic in 19 out of 20 cases and improved upon the average heuristic performance by approximately 3\% in terms of makespan.

These findings demonstrate that our method not only surpasses the overall average of heuristic solutions but also remains competitive, even superior, when compared against a super heuristic that selects the best heuristic per instance.

\begin{figure}[ht]
\centering
\includegraphics[width=0.8\linewidth]{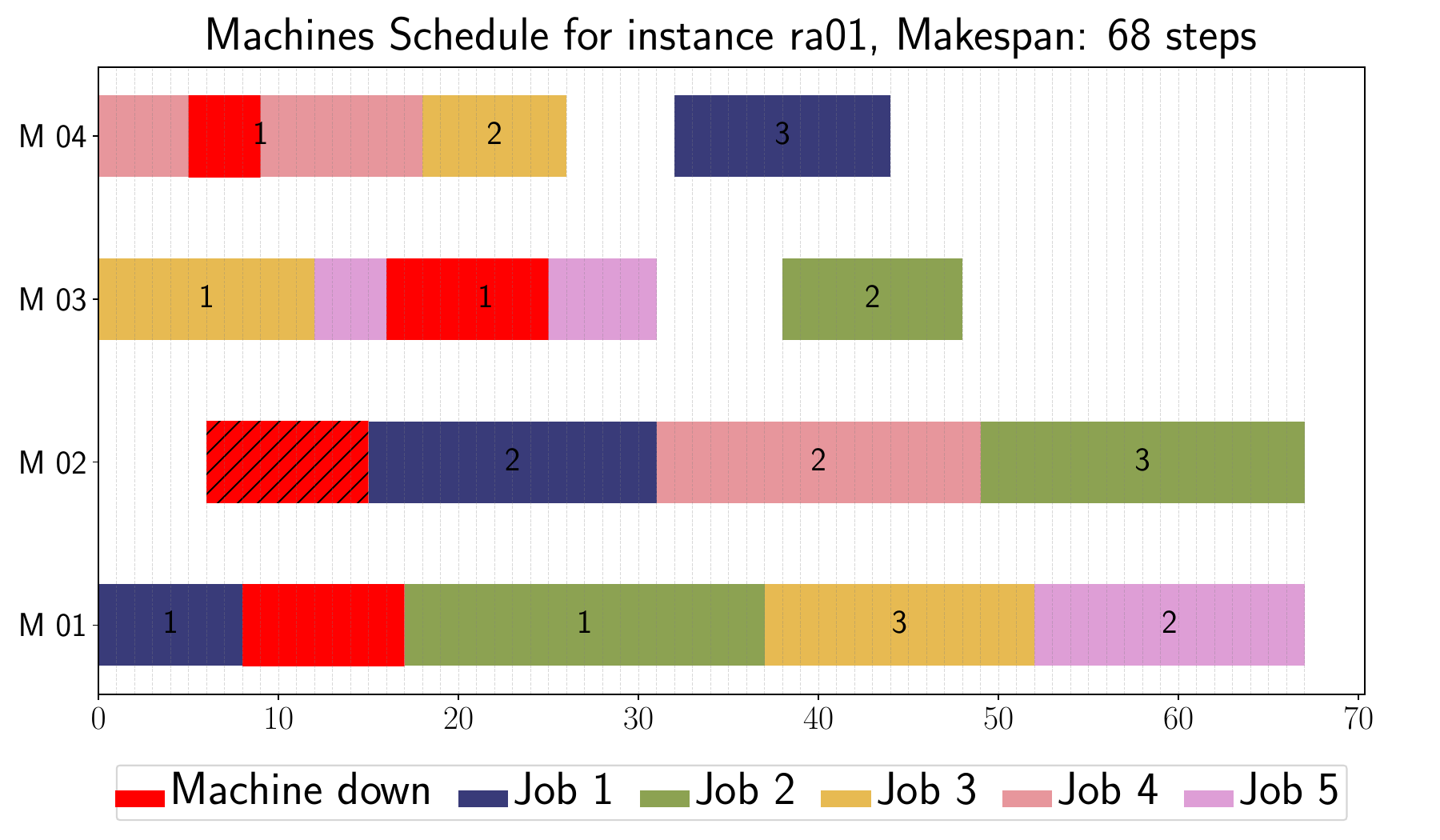}
\caption{\textbf{Gantt chart of the RL agent-generated schedule for an instance with five jobs and four machines (Raj 01).}}
\label{fig: Gantt chart model}
\end{figure}

To visually inspect the RL-generated schedule, we show in Fig.~\ref{fig: Gantt chart model} an example of the Gantt chart. This visualization highlights three distinct machine unavailability patterns occurring during job execution, before job start, and during machine idle time.

First, machine unavailability occurs in the middle of job execution. This is evident on Machine 4 at time step 6 for Job 4, and on Machine 3 at time step 17 for Job 5. These interruptions simulate machine breakdowns occurring during processing. In such cases, the operation is temporarily halted. For instance, Operation 1 of Job 4 on Machine 4 took a total of 18 time steps to complete, including 14 steps of actual processing dictated by the instance and four steps of downtime.

Second, unavailability may occur between operations, preventing the immediate start of the next job. This can be observed on Machine 1 at time step 9, where, after completing Operation 1 of Job 1, Operation 1 of Job 2 does not start until the machine becomes available again. This scenario can represent required tool changes or setup times between operations.

Finally, a machine can become unavailable while idle, as seen on Machine 2 at time step 7. This model planned maintenance or preventive servicing during idle periods before starting a new operation.

\begin{table*}[ht]
\centering
\caption{Comparison of our method with dispatching heuristics on various Taillard instances with machine breakdown scenario}
\resizebox{\textwidth}{!}{%
\begin{tabular}{lccccccccccccccc}
\toprule
Size & FIFO & SPT & LPT & SPS & LPS & LTWR & MTWR & SPSR & LPSR & SPTN & LPTN & LWT & \makecell{AVG\\Heur.} & Ours & \makecell{Gap\\\%} \\ 
\midrule
15$\times$15 & 1587 & 1568 & 1578 & 1587 & 1587 & 1567 & 1575 & 1584 & 1588 & 1562 & 1588 & 1587 & 1580 & \textbf{1545} & \textbf{2.2\%} \\
20$\times$15 & 1807 & 1852 & 1858 & 1807 & 1807 & 1851 & 1861 & 1803 & 1803 & 1856 & 1829 & 1807 & 1829 & \textbf{1797} & \textbf{1.7\%} \\
20$\times$20 & 2121 & 2189 & 2123 & 2121 & 2121 & 2199 & 2118 & 2123 & 2115 & 2214 & 2120 & 2121 & 2140 & \textbf{2114} & \textbf{1.2\%} \\
30$\times$15 & 2304 & 2303 & 2322 & 2304 & 2304 & 2306 & 2323 & 2301 & 2306 & 2311 & 2304 & 2304 & 2308 & \textbf{2296} & \textbf{0.5\%} \\
30$\times$20 & 2711 & 2696 & 2746 & 2711 & 2711 & 2695 & 2752 & 2717 & 2706 & 2742 & 2705 & 2711 & 2717 & \textbf{2691} & \textbf{1.0\%} \\
50$\times$15 & 3699 & 3594 & 3664 & 3699 & 3699 & 3594 & 3665 & 3694 & 3696 & 3575 & 3629 & 3699 & 3659 & \textbf{3627} & \textbf{0.9\%} \\
50$\times$20 & 3678 & 3699 & 3731 & 3678 & 3678 & 3704 & 3736 & 3682 & 3677 & 3684 & 3690 & 3678 & 3693 & \textbf{3651} & \textbf{1.1\%} \\
100$\times$20 & 6297 & 6433 & 6310 & 6297 & 6297 & 6433 & 6310 & 6295 & 6294 & 6268 & 6327 & 6297 & 6321 & \textbf{6297} & \textbf{0.4\%} \\
\hline
\textbf{Average} & 3025 & 3042 & 3041 & 3025 & 3025 & 3044 & 3042 & 3025 & 3023 & 3027 & 3024 & 3025 & 3031 & \textbf{3002} & \textbf{1.1\%} \\
\bottomrule
\end{tabular}}
\label{tab:heuristic_comparison}
\end{table*}

After validating our approach on small-sized instances, we extended our evaluation to larger Taillard instances ranging from \(15 \times 15\) to \(100 \times 20\) (jobs \(\times\) machines). The results, presented in Table~\ref{tab:heuristic_comparison}, report the mean makespan over 100 independent runs. Our method consistently outperforms the average of heuristics across all instance sizes, with relative improvements ranging from 0.4\% to 2.2\%. On average, our approach achieves a makespan reduction of approximately 30 time steps compared to the average of the heuristics. Furthermore, a 95\% confidence interval of around 27 steps confirms that our method reliably surpasses heuristic baselines, even in the presence of randomness.

\subsection{Results under Machine breakdown and random job arrival}
\label{subsec: Results under Machine breakdown and random job arrival}

\begin{figure}[ht]
\centering
\includegraphics[width=0.8\linewidth]{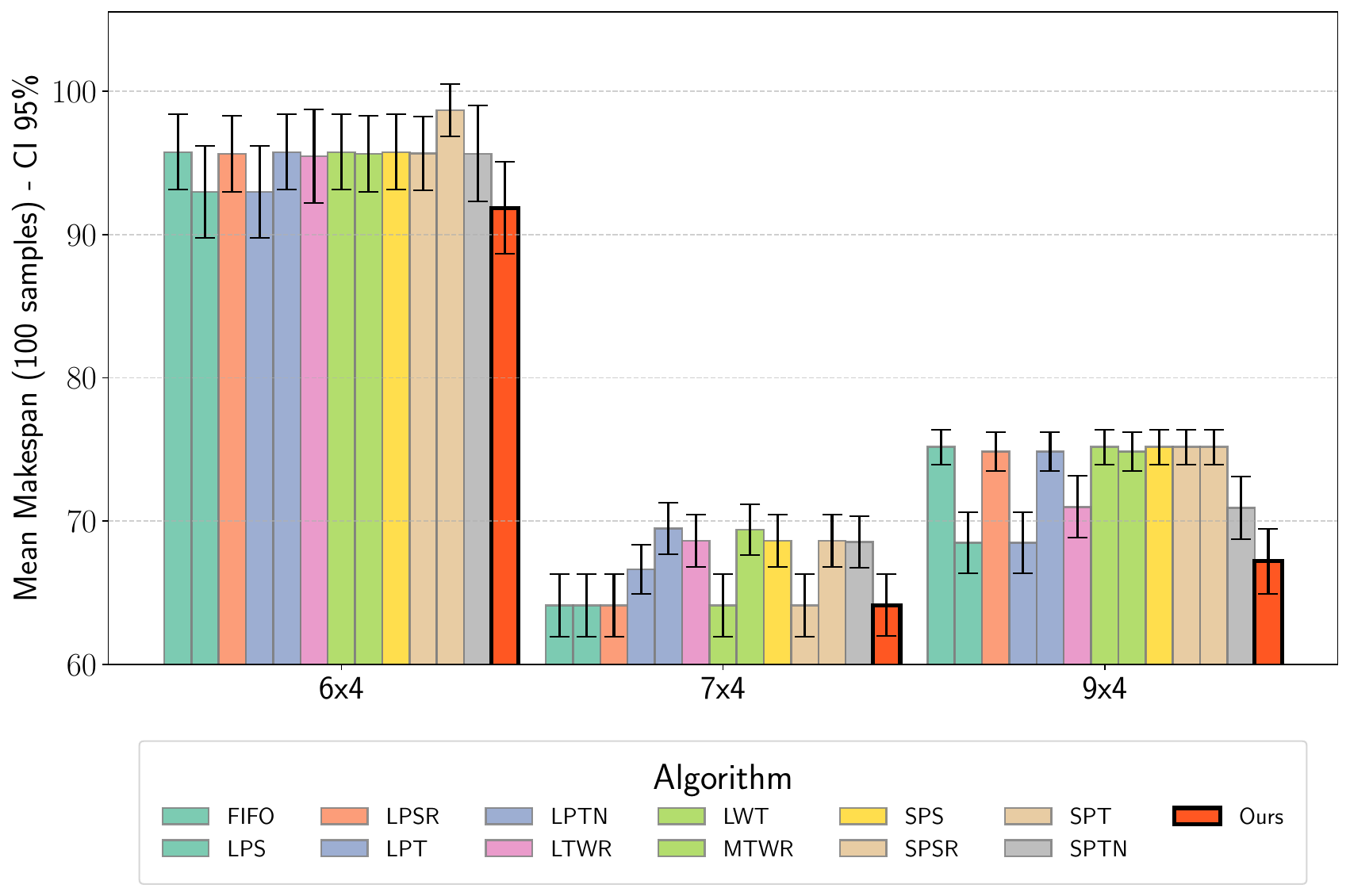}
\caption{Comparison of RL agent performance against various heuristics on instance sizes (6×4), (7×4), and (9×4) (Jobs × Machines) under simultaneous random job arrivals and machine breakdown scenarios. The results show the average makespan over 100 runs along with the 95\% confidence interval.}
\label{fig: small instances results}
\end{figure}

In Fig.~\ref{fig: small instances results}, we present a comparative analysis of our approach against a range of heuristics under the combined presence of machine breakdowns and random job arrivals, evaluated across three instance sizes. The bar chart represents the mean makespan for 100 runs and the 95\% confidence interval. In this setting, job operations are not fully available at the start; instead, tokens are progressively released from the planned job place to the job place, as described in Section~\ref{subsec: Modelling the random job arrivals}. The results show that our method achieves an average improvement of approximately 2\% over the heuristics, with gains up to 8\% depending on the instance. The average width of the 95\% confidence interval is 2.18 steps higher than in the machine-breakdown-only scenario, which can be attributed to the additional uncertainty introduced by the random job arrivals.

\subsection{Ablation Study}
\label{subsec: Ablation Study}

This section presents an ablation study focused on dynamic masking as a key component for handling dynamic events.

Invalid actions can be addressed in two ways: either by proactively preventing them before execution or by allowing the agent to select freely and subsequently handling any invalid choices through reactive correction mechanisms.  In our setup, we apply masking upfront by modifying the policy network’s logits. Based on the Petri net's guard functions, invalid actions receive a large negative value, making their softmax probabilities effectively zero and preventing their selection.

To evaluate the effect of masking, we allow the agent to choose any action, including invalid ones. If an invalid action is selected, the environment ignores it, leaving the state unchanged. While simple, this approach has a key drawback: in a deterministic environment, the same state leads to the same action, potentially causing the agent to get stuck in an infinite loop. To prevent this, a valid action is randomly selected whenever an invalid one is chosen, ensuring progress.

It is also possible to assign a penalty when the agent selects an invalid action. However, in this ablation, we choose to ignore invalid actions instead, for two main reasons. First, since the collected reward is used as a performance metric, penalizing invalid actions would artificially lower the total reward and make the masked agent appear to perform better. Second, penalization shifts the learning focus toward learning the masking function instead of decision-making.
 
\begin{figure}[H]
\centering
\includegraphics[width=0.8\linewidth]{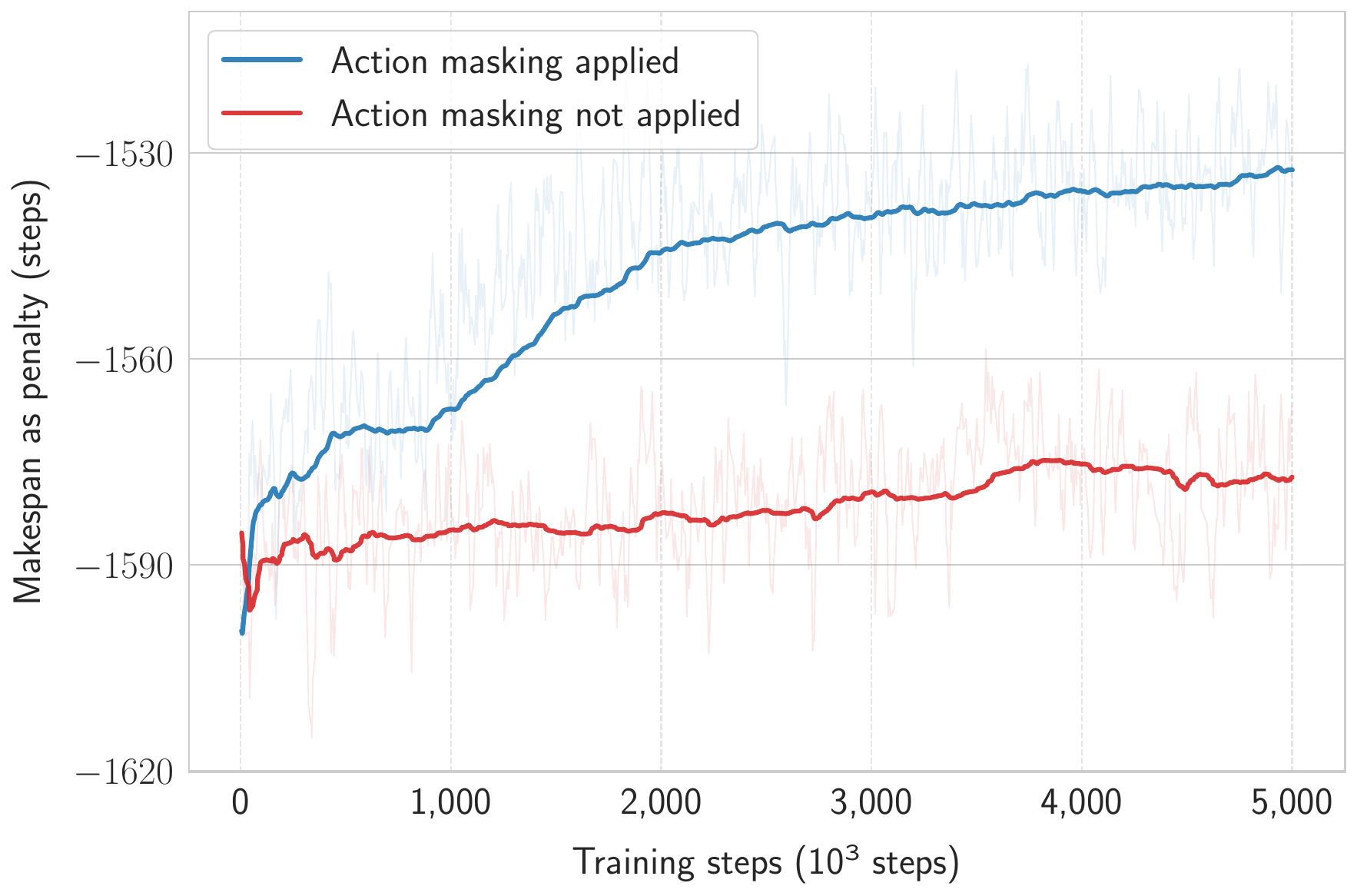}
\caption{Comparison of Collected Reward Evolution Over Training Episodes for Agents With and Without Action Masking}
\label{fig: ablation}
\end{figure}

Figure~\ref{fig: ablation} shows the training reward progression of the agent on a 15-job, 15-machine JSSP instance over $5 \times 10^6$ time steps under a machine breakdown scenario. With masking, the collected reward increases steadily, indicating more stable learning and improved credit assignment. This suggests that masking effectively guides the agent toward valid decisions and accelerates convergence to a better policy. In contrast, training without masking exhibits less stability, with frequent fluctuations and a lower final reward, highlighting the difficulty of learning in the presence of invalid actions and the added value of masking.

\section{Conclusion and Future Work}
\label{sec: Conclusion}

In this paper, we demonstrate that our approach, "PetriRL," can effectively handle various types of disruptions in DJSSP, including random job arrivals and machine unavailability. These dynamic events were embedded within the simulation environment modeled using a CPTN, where machine availability followed a Weibull distribution to capture aging effects, and job arrivals were generated using a Gamma distribution to reflect realistic patterns such as clustering and burstiness. We employed a Maskable PPO algorithm as the RL agent, with action masking derived from the Petri net's guard functions based on the current token distribution. This enabled the RL agent to consider only valid actions and utilize the available samples efficiently.

We evaluated our approach against standard dispatching rules under scenarios involving machine breakdowns as well as the simultaneous occurrence of breakdowns and random job arrivals. Experiments on both small-scale problems and large instances from the Taillard benchmark showed that our method consistently outperformed traditional rules. On average, our RL-based approach improved performance by 3\% on small instances and by 1.1\% on large instances when compared to the average performance of heuristic baselines. The improvement was particularly notable on larger instances due to longer decision sequences, allowing the RL policy to demonstrate its ability to adapt.

In summary, model-based reinforcement learning, when combined with interpretable Petri net models, provides a promising and scalable alternative to heuristic methods for real-time, disruption-resilient scheduling. A promising direction for future work is to incorporate prior distribution characteristics of random events into the agent's observations. This may help the policy learn more anticipatory strategies and improve robustness under uncertainty.

\section*{Statements and Declarations}

\subsection*{Funding}

This publication is funded by the Open Access Publication Fund of South Westphalia University of Applied Sciences.

\subsection*{Competing Interests}
The authors have no relevant financial or non-financial interests to disclose.

\subsection*{Data availability statement}

The Python package developed for this study is openly available on 
\href{https://pypi.org/project/petrirl/}{PyPI}. 
The scripts and experimental results are available on 
\href{https://github.com/Sofiene-Uni/PetriRL_DJSSP}{GitHub}.

\bibliographystyle{elsarticle-num} 
\bibliography{References.bib}


\end{document}